\documentclass{article}

\pdfoutput=1

\usepackage[preprint]{neurips_2019}
\usepackage[utf8]{inputenc} 
\usepackage{hyperref}       
\usepackage{url}
\usepackage{amsfonts,bm}      
\usepackage{microtype}      
\usepackage{booktabs}
\usepackage{amsmath}
\usepackage{multirow}
\usepackage{tcolorbox}
\newtcolorbox{example}[1]{colback=blue!5!white, colframe=blue!20!black!60!white,fonttitle=\bfseries,title=#1}
\setcitestyle{numbers}
\setcitestyle{square}
\setcitestyle{comma}

\title{Neural Theorem Provers Do Not Learn Rules Without Exploration}

\author{Michiel de Jong\\
  University of Southern California\\
  Los Angeles, CA 90089\\
  \texttt{msdejong@usc.edu} \\
   \And Fei Sha \\
   Google AI \footnotemark \\
   Los Angeles, CA 90089 \\
   \texttt{fsha@google.com} \\
}

\begin{document}
\maketitle
\footnotetext{On leave from University of Southern California (feisha@usc.edu)}

\begin{abstract}
Neural symbolic processing aims to combine the generalization of logical learning approaches and the performance of neural networks. The Neural Theorem Proving (NTP) model by Rocktaschel et al (2017) learns embeddings for concepts and performs logical unification. While NTP is promising and effective in predicting facts accurately, we have little knowledge how well it can extract true relationship among data. To this end, we create synthetic logical datasets with injected relationships, which can be generated on-the-fly, to test neural-based relation learning algorithms including NTP.  We show that it has difficulty recovering relationships in all but the simplest settings.  Critical analysis and diagnostic experiments suggest that the optimization algorithm suffers from poor local minima due to its greedy winner-takes-all strategy in identifying the most informative structure (proof path) to pursue. We alter the NTP algorithm to increase exploration, which sharply improves performance.  We argue and demonstate that it is insightful to benchmark with synthetic data with ground-truth relationships, for both evaluating models and revealing algorithmic issues. 
\end{abstract}

% !TEX root = main.tex
\section{Introduction}

Neural networks have advanced the state of the art on a variety of prediction tasks.
However, they are also data-hungry, transfer poorly to data generated from different
distributions~\cite{barrett2018measuring, lake2017generalization}, and their decision-making is difficult to
interpret~\cite{chakraborty2017interpretability}. On the other hand, learning logical rules through Inductive Logic Programming~\cite{muggleton1994inductive, muggleton2012ilp} has good
generalization properties, but does not cope well with noisy data or fuzzy
relationships~\cite{de2008probabilistic}, and is difficult to scale beyond toy datasets because of the large search
space~\cite{zeng2014quickfoil}. Recently, there have been renewed efforts to combine the two paradigms by applying 
deep learning methods to classical symbolic AI methods.

Several differentiable logic learning approaches have been proposed~\cite{rocktaschel2017end, evans2018learning,
yang2017differentiable} that are better equipped to deal with noisy data, and employ gradient descent rather than
discrete search to cope with a large search space. \citet{rocktaschel2017end} proposes Neural Theorem Prover (NTP) to fuse subsymbolic methods and logical
rule learning by endowing constants and predicates with vector embeddings.
We view this as a  promising direction, as the embedding approach lets the model learn the
semantics of entities and concepts concurrently with the relationships between those concepts, as well as allowing
for nuanced and ambiguous rules. However, as with other differentiable logic models, this model currently only
works on limited toy datasets. In order to make progress we need to challenge the model and understand its
limitations as well as how those limitations can be addressed. 

NTP and its variants have inspired a lot of interests. Nonetheless, one of its most important motivations ---  learning logical rules improves generalization --- has not been critically analyzed.  In particular, \emph{how well does NTP learn rules?}  Like other models, NTP was primarily evaluated on its fact prediction accuracy, not rule learning performance as ground-truth rules in real data are hard to identify. 

To address this gap, we create synthetic logical datasets with injected relationships (as ground-truths) and use them to measure directly rule learning performance. While we focus on NTP in this paper, the datasets can be used to assess other learning algorithms.

Our findings of NTP are  surprising: NTP does not learn rules well, except when the relationship is very simple,
even though it can achieve high fact prediction accuracy.  Diagnostic experiments suggest learning the model suffers from
poor optimization. In particular, a wrong proof for a true fact can start out with a high endorsing score, due to
initialization. However, the optimization will continue to greedily increase the score of the wrong proof.  This
winner-takes-call credit assignment leads to  systematically undesirable local minima, leading to incorrect
relationships being learned.

Fortunately, this weakness can be easily remedied. We take inspiration from beam search, a common trick in speech and language processing. We retain a subset of proofs even they are scored low initially. This allows optimization to explore other alternative proof paths and eventually leads to a very large increase in performance of rule learning.

The main contribution of this work is to single out the need of measuring rule learning performance in addition to fact prediction accuracy. We believe it is imperative to have challenging benchmark datasets while such performance can be measured under different scenarios. To this end, we have curated a recipe to generate multiple datasets with varying degrees of complexity (in terms of underlying sets of rules). We analyze the root cause of NTP failing to achieve high performance on rule learning and propose a simple fix to optimization, thus demonstrating the utility of applying those datasets to advancing research in neural subsymbolic processing.

% !TEX root = ms.tex
\section{Related work}

In a broad sense, our work aims to learn relationships between logical predicates. There exists a very large body
of literature that shares this objective. Inductive Logic Programming (ILP) employs search methods to find logical
rules that best explain the data~\cite{muggleton1994inductive, muggleton2012ilp, de2008probabilistic,
zeng2014quickfoil}. Statistical relationship learning (SRL) also looks to learn relationships between logical predicates, but generally
employs statistical methods to learn more general probabilistic relationships, not necessarily in
the form of logical rules~\cite{koller2007introduction, getoor2007probabilistic, sutton1}. Some of the more recent
work in this vein uses methods from or inspired by deep learning. \citet{trouillon2016complex} assigns an embedding
to each logical constant and predicate, and predicts the presence of a predicate relationship between constants by
the triplet product of the embeddings. Graph neural networks~\cite{wu2019comprehensive} operate over a graph
structure, which is inherently a relational setting. \citet{dong2019logic} trains a deep neural network to predict
a probability vector of output predicates from a probability vector of input predicates, naturally extending SRL to
deep learning. 

Differentiable logic learners employ some form of continuous relaxation of the logical reasoning process so that
its parameters can be trained through gradient descent \cite{evans2018learning, serafini2016logic,
manhaeve2018deepproblog, rocktaschel2017end}. One family of models generates logical clauses and associates
predicates with a trained attention vector over these clauses \cite{evans2018learning, yang2017differentiable}. In
contrast, neural theorem proving (NTP), which is the focus of this paper, endows predicates, constants and rules
with embeddings that are trained through gradient descent, and receive meaning by their position in the embedding
space~\cite{rocktaschel2017end, minervini2018towards, campero2018logical}. The embedding approach allows the model
to learn semantics of entities and concepts, in a similar way to word embeddings in
NLP~\cite{minervini2018towards}. 

We contribute to this line of work by systematically generating synthetic data to directly measure rule-learning
performance. Existing work has either used very small toy examples~\cite{evans2018learning} or real data without
known ground-truth relationships. Although the problem of local minima when learning discrete structures is well known, to our knowledge we are the first work to closely study how this
effect impacts performance and show that increased exploration can address the problem in the differentiable logic
setting.

% !TEX root = ms.tex
\section{Neural-based inductive logic programming}
\label{basesection}

\subsection{Problem statement} 

NTP operates on a subset of first-order logic. Constants $\textsc{c} \in \mathcal{C}$ represent entities, such as
individuals or objects. Predicates $\texttt{P}: \mathcal{C}^k \rightarrow \{0, 1\}$ are boolean functions of sets
of constants, which represent properties of or relationships between entities. For example,  
\texttt{GrandparentOf}$(x, y)$ is a predicate, taking value 1 if $x$ is the grandparent of $y$ and $0$ otherwise. The number
of arguments in a predicate is called its order. Logical datasets consist of facts, which are statements that a
predicate holds for a particular set of constants, for example, \texttt{GrandparentOf}$(\textsc {\textsc{Harry}},
\textsc{\textsc{Emily}})$. We also have variables and rules. Variables $v$ are symbols that may represent any
constant. We consider rules of the form $\texttt{P}_0(v_0) \leftarrow \texttt{P}_1 (v_1) \land \cdots
\texttt{P}_{\textsc{R}} (v_\textsc{R})$, where $v_i$ are sets of variables (possibly overlapping). Collectively,
all $\texttt{P}_r(v_r)$ are called the body and $\texttt{P}_0(v_0)$ is called the head of the rule. Such a rule
implies that, for any assignment of constants to $v_0, v_1, \cdots, v_\textsc{R}$, if all the predicates in the body hold,
then the head also holds. $\textsc{R}$ is called the size of the rule.

We consider two types of learning settings. In \emph{fact prediction} task~\cite{rocktaschel2017end}, we are interested in predicting the
truth value of a set of test facts after learning from the true facts in the training data $D$. %This is the learning setting primarily used in.

The other type of learning focuses on \emph{relation learning}, where we are interested in identifying rules from
the training dataset. Once identified, we can apply these rules to predict the truth of unknown test facts. Note
that this type of learning may be more challenging, but potentially much more beneficial in many application
settings, for instance, learning with a small set of training instances.

\subsection{Neural Theorem Proving}

The basic idea in NTP is to embed symbols such as predicates and constants from the the data as embeddings $\theta \in
\mathcal{R}^d$. Rules are represented by tuples of predicate embeddings. Inference is the process of applying facts
and rules to derive conclusions. Since symbols are all points in the embedding space, 
``applying"  means computing minimum distances among them. Learning happens by placing the embeddings
in a position such that correct chains of facts and rules that prove valid facts have minimum distance. A more detailed description can be found
in~\cite{rocktaschel2017end}.

\textbf{Inference}  Given a set of facts and rules, logical programming infers the truth value of a goal fact by
attempting to prove that goal the fact is derivable from other facts and rules. NTP is loosely based on using \emph{unification} in a
backward chaining proving technique~\cite{russell2016artificial}. 

In standard logical inference, unification compares logical statements and makes any necessary variable
substitutions to make those statements identical. For example, in order to unify
\texttt{GrandparentOf}(\textsc{Harry}, \textsc{Emily}) and \texttt{GrandparentOf}(\textsc{Harry}, $x$), we make the
substitution $x = \textsc{Emily}$. On the other hand, \texttt{GrandparentOf}(\textsc{Alex}, \textsc{Emily}) and
\texttt{GrandparentOf}(\textsc{Harry}, $x$) cannot be unified as the constant \textsc{Alex} is not the same as
\textsc{Harry}. \texttt{GrandparentOf}(\textsc{Harry}, \textsc{Emily}) and \texttt{ParentOf}(\textsc{Harry}, $x$)
cannot be unified either as the predicates do not match.

Standard backward chaining attempts to unify the goal fact with training facts as well as the head of rules. If the
goal fact unifies with any training fact, the proof has been completed. Otherwise, if the fact matches a rule head,
the rule body predicates are added to a queue to be proved. This process continues recursively until all necessary
facts in the queue are successfully proved, in which case the goal fact has been proved, or until all facts and
rules have been tried without finding a proof. 

For NTP, unification and backward chaining work somewhat differently. A fact can be unified with another fact
or rule head even if the predicate or constants in the fact are different, as long as the predicate has the same
order. If predicates or constants differ, the unification is given a score that is a function of the distance
between the embeddings of the symbols being unified. The score of a proof is a function of the
embedding distance of the ``worst'' unification in the proof. Let $U$ be the set of unifications in
the proof, and $S_u$ and $V_u$ the logical symbols in unification $u$, then the proof score $\rho$ is
\begin{equation}
	\label{proofscore}
	\rho (U) = \min_{u \in U} \exp(-||\theta_{S_u} - \theta_{V_u}||_2)  
\end{equation}
An example of a simple proof of \texttt{HasSibling}(\textsc{Emily}), given that we know
\texttt{HasBrother}(\textsc{Emily}), would be to directly unify \texttt{HasSibling}(\textsc{Emily}) and
\texttt{HasBrother}(\textsc{Emily}). The score of this proof would be $\exp(-||\theta_{\texttt{HasSibling}} -
\theta_{\texttt{HasBrother}}||_2)$.

In this definition of unification, there can be many successful proofs. We calculate
the scores of all possible proofs, and define the score of a fact to be equal to the score of the best proof of
that fact. The learning process involves raising the proof score for known facts as much as possible.

\textbf{Learning} Rules are instantiated from rule templates, such as $\texttt{C}(x, y) \leftarrow \texttt{A}(x, y) \land \texttt{B}(x,
y)$. Rules are created by instantiating an embedding for each predicate in the template; in this
template we would generate embeddings $\theta_\texttt{A}$, $\theta_\texttt{B}$ and $\theta_\texttt{C}$. Unlike the data predicates, these rule
predicates do not have any intrinsic meaning, but take on a meaning during training by being placed close in the
embedding space to data predicates. All symbols' embeddings are randomly instantiated. 

Models are trained by attempting to prove logical facts. The goal during training is
to promote rules that successfully prove true facts (facts present in the training set) and discourage rules that
successfully prove false facts (generated by corrupting facts in the training set). The NTP algorithm assigns a
score to each fact equal to the score of the highest scoring proof of that fact. Let $\bm{\theta}$ be the set of
all embeddings. Given a truth value $y_i$ and a fact score $\rho_i = \max_{U} \rho(U)$, the loss is given by
\begin{equation}
	\label{eqn:loss}
	\mathcal{L}_i(\bm{\theta}) = -y_i \log(\rho_i) - (1 - y_i) \log (1 - \rho_i )
\end{equation}
Note that the score of the fact, while bounded between 0 and 1, should not be interpreted as a normalized
probability. Nonetheless, the loss is differentiable w.r.t. $\bm{\theta}$. In a single training step, the
two embeddings in the \textit{worst} unification of the \textit{best} proof of the training fact are updated. See below for an illustrative example:

\begin{example}{Example I: Learning procedure}
\small 
	Consider a task that requires learning the relationship $\texttt{HasSibling}(x) \leftarrow
	\texttt{HasBrother}(x)$. A rule is instantiated from a corresponding rule template $B \leftarrow A$. Embeddings
	are initialized for \texttt{HasSibling}, \texttt{HasBrother}, $A$, $B$, as well as any constants and other
	predicates in the task. \\[0.5em]
	The first training data point is \texttt{HasSibling}(\textsc{Emily}). \texttt{HasBrother}(\textsc{Emily}) is
	also in the training dataset. The best proof for \texttt{HasSibling}(\textsc{Emily}) unifies
	\texttt{HasBrother} with $B$ and \texttt{HasSibling} with $A$ and applies the rule to prove
	\texttt{HasSibling}(\textsc{Emily}) from \texttt{HasBrother}(\textsc{Emily}). $A$ and \texttt{HasSibling} were
	the furthest apart, and since the proof has successfully led to a true fact the next gradient step brings them
	closer together. Alongside \texttt{HasSibling}(\textsc{Emily}), we also train on the corruption
	\texttt{HasSibling}(\textsc{Alex}), which is not a fact in the training data. \texttt{HasChild}(\textsc{Alex})
	is also in the training data, and the best proof for \texttt{HasSibling}(\textsc{Alex}) ends up unifying
	(\texttt{HasSibling}, \texttt{HasChild}). Since this proof leads to an untrue fact, the next gradient step
	places \texttt{HasSibling} and \texttt{HasChild} further apart.
\end{example}

% !TEX root = ms.tex
\section{Design synthetic datasets for evaluating NTP}

\subsection{A Recipe for data generation}

We use the following recipe to procedurally generate datasets. Our main design consideration is to finely control
the properties of our data, so that it is possible to investigate the effect of different factors in isolation.
First, we choose the size and the order of the predicates to be used. Next, we select the number of constants $n_c$
and predicates $n_p$. Each predicate has a base truth probability of $p_b$ which is identical for all inputs $x \in
\mathcal{C}^k$. $n_{rel}$ relationships between predicates are then injected from a relationship template. In those (ground-truth) relationships, for a given input $x$, if all the predicates
$P_r(x)$ in the body are true, then the predicate in the head has an increased probability $p_r > p_b$ of being
true.

Using this generic recipe, we flexibly generate datasets on-the-fly for each experiment. Table~\ref{expparams} outlines the
parameters of the data generation process used for our main experiments, depending on the body of the relationship.
Experiments with binary predicates use fewer constants and a lower base probability to reduce computational
cost, given that binary facts scale quadratically with constants. 

\begin{table}[t]
	\caption{Various configurations for generating synthetic datasets on-the-fly}
	\label{expparams}
	\centering
	\begin{tabular}{c|c|c|c|c|c|c|c}		
	\hline			
		\multicolumn{2}{c|}{Relationship body} & Rule &\multicolumn{5}{c}{Parameters} \\
		\cline{1-2} \cline{4-8}
		size & order  & template & $n_c$ & $n_p$ & $p_b$ & $p_r$ & $n_{rel}$ \\
		\hline
		\multirow{2}{*}{1}    & unary  & $\texttt{P}_2(x) \leftarrow \texttt{P}_1(x)$ & 200   & \multirow{5}{*}{5}
		& 0.5 & \multirow{5}{*}{1.0} & \multirow{5}{*}{1} \\
		    & binary & $\texttt{P}_2(x, y) \leftarrow \texttt{P}_1(x, y)$ & 60   &    & 0.25 &   &  \\ 
		    \cline{1-4} \cline{6-6} 
		\multirow{2}{*}{2}    & unary & $\texttt{P}_0(x) \leftarrow \texttt{P}_1(x) \land \texttt{P}_2(x)$   & 400
		&    & 0.5 &    & \\
		    & binary  & $\texttt{P}_0(x,y)\leftarrow \texttt{P}_1(x, y) \land \texttt{P}_2(x, y)$ & 60   &      &
		    0.25 &   &  \\
		    \cline{1-4} \cline{6-6} 
		3    & unary & $\texttt{P}_0(x) \leftarrow \texttt{P}_1(x) \land \texttt{P}_2(x) \land \texttt{P}_3(x)$   &
		800  &   & 0.5 & & \\
		\hline
	\end{tabular}
\end{table}

\subsection{Evaluation}

The NTP algorithm learns embeddings for $n_t$ rule instantiations. Each predicate embedding of an instantiation is decoded to the data predicate that is closest to it in the embedding space. The overall rule
decoding is equal to the score of the worst predicate decoding, calculated as in
Eqn~\ref{proofscore}. 

Relationship learning performance is measured by recall and precision, measured and averaged over a number of runs. Recall for a single run is defined as the proportion of injected relationships for which at least one rule decodes to that relationship. We then take the mean over all runs. For precision measure, we construct a list of all decoding scores, and a corresponding list of ``gold values", that equal 1 if the corresponding rule decoding matches an injected relationship and 0 otherwise. The precision measure (PR-AUC) is then defined as the area under the PR for these lists.

Fact prediction accuracy is normally evaluated out on a randomly held-out test set. For NTP, fact evaluation is
complicated by the fact that the algorithm relies on constant embeddings that are learned during training. 
Moreover, since predicate truth values not involved in a relationship are generated i.i.d, the model cannot learn to predict these, and performance on these facts is not very informative. We instead split off a small proportion of \textit{active facts}, defined as facts which could be predicted to be true if the relationship is learned successfully. 

We corrupt each test fact by changing the constants in all possible ways such that the result is not present in the training data, and apply the
model to the facts and corruptions. The first evaluation measure is the mean reciprocal rank (MRR) of the true fact relative to corrupted facts. This measure depends on the number of corruptions, which in turn depends on
the size of the dataset, making it hard to compare between tasks. Therefore, we also compute a size-invariant
measure by duplicating the score of the true fact by number of corrupted facts, and
calculating the area under the ROC curve, with target 1 for true facts and 0 for corruptions. 

% !TEX root = ms.tex
\section{Critical analysis of NTP}

In this section, we perform a critical analysis of the NTP model with the synthetic datasets we have created. Prior
work has shown the method attains good test accuracy in predicting truth values of facts~\cite{rocktaschel2017end}.
We analyze the model's ability to identify rules under different conditions, revealing that the base NTP model has
poor rule learning performance, and investigate why this is the case. 

\subsection{Additional experiment details}

During the training, each batch contains 10 true
facts from the training set. For each true fact, there is also one corrupted fact per predicate argument (so 1 for
unary predicates, 2 for binary predicates) which is generated by randomly selecting a constant for which the
predicate does not hold in the training set. For each of the 50 runs,  3 (randomly initialized) rules are instantiated from the
correct template. The hope is at least one of them will be driven close to the correct relationship. The model is trained for 50 epochs using the Adam optimizer~\cite{kingma2014adam}
with a learning rate of $10^{-3}$, gradient clipping of (-5, 5) and exponential learning rate decay of $3\times 10^{-4}$. 

We adhere to the evaluation protocol set in the previous section. None of the parameter choices ($n_c$, $n_p$,
$p_b$, $p_r$ or $n_{rel}$) qualitatively affect the results in this paper (the supplemental material presents
detailed experiment results). The code can be found at \url{https://github.com/Michiel29/ntp-release}.

\subsection{Ablation studies}
\label{ablations}
\begin{table}[t]
	\caption{NTP performance by number of constants }
	\label{constants}
	\centering
	{\small 
	\begin{tabular}{cccccccc}
						 
		\multicolumn{2}{c}{Rule body} & \multicolumn{3}{c}{Data size}
		&\multicolumn{2}{c}{Rule performance} & Fact Performance \\
		\cmidrule(lr){1-2}
		\cmidrule(lr){3-5}
		\cmidrule(lr){6-7}
		\cmidrule(lr){8-8}
		Size & Order & Constants & Total facts & Active facts & Recall & PR-AUC  & ROC-AUC \\
		\midrule
		\multirow{4}{*}{1}& \multirow{4}{*}{Unary} &  50  &  137  & 23 & 0.42 & 0.51 & 0.80  \\
		&& 100 & 274 & 47 & 0.46 & 0.66 & 0.88 \\
		&& 200 & 546 & 92 & 0.6 & 0.76 & 0.92  \\
		&& 800  & 2194 & 388 & 0.46 & 0.68 & 0.96 \\
		\midrule
		\multirow{4}{*}{2}& \multirow{4}{*}{Unary} &  50  & 133 & 11 & 0.0 & 0.31 & 0.71     \\
		&& 100 & 266 & 21 & 0.0 & 0.36 & 0.75 \\
		&& 200 & 532 & 43 & 0.02 & 0.37 & 0.78  \\
		&& 800 & 2139 & 185& 0.02 & 0.4 & 0.81 \\    
		\bottomrule
	\end{tabular}
	} \vskip -1em
\end{table}

\paragraph{How much data does NTP need?}  One of the appealing properties of rule-learning algorithms is that they tend to be data efficient. Can NTP achieve this goal?  Table~\ref{constants} shows relationship learning and fact prediction accuracy as a function of the number of constants in the data. It also shows the resulting total facts and active facts, the latter of which constitute the actual signal in the data. We restrict ourselves to unary predicates for this experiment, as scaling binary predicates in this way becomes computationally expensive. 

When NTP learns anything at all, it seems to be able to do so from very little data. Note that, while performance increases modestly with data for body size one, performance stays low at body size two. We discuss the likely cause this poor performance in the next section.

\begin{table}[t]
	\caption{NTP performance by relationship type and the number of predicates}
	\label{reltype}
	\centering
	{\small
	\begin{tabular}{ccccccc}
						 
		\multicolumn{2}{c}{Rule body} & \multirow{2}{*}{$n_p$} & \multicolumn{2}{c}{Rule performance}  & \multicolumn{2}{c}{Fact performance}
		\\
		\cmidrule(lr){1-2}
		\cmidrule(lr){4-5}
		\cmidrule(lr){6-7}
		Size & Order &  & Recall & PR-AUC & MRR  & ROC-AUC \\
		\midrule
		1 & unary  & \multirow{5}{*}{5} & 0.48   & 0.62   & 0.27 & 0.91    \\
		1    & binary &  & 0.58   & 0.74   & 0.57 & 0.98    \\  
		2    & unary  &   &0.04   & 0.37   & 0.04 & 0.81    \\
		2    & binary &  & 0.02   & 0.4    & 0.15 & 0.89    \\
		3    & unary  &  & 0.02  & 0.4 & 0.02 & 0.81\\
		\hline
		\hline
		\multirow{2}{*}{1}    & \multirow{2}{*}{unary}	 & 3 & 0.58 & 0.76 & 0.29 & 0.95 \\
		& & 10 & 0.34 & 0.58 & 0.23 & 0.87  \\
		\hline

	\end{tabular}
	} \vskip -1em
\end{table}

\paragraph{Can NTP learn complex relationships?} An important desiderata for relation learning is to scale to learn complex relationships. The top half of the Table~\ref{reltype} shows relationship learning and fact prediction accuracy by NTP under different sizes and orders of the relationships.  The performances are reasonable for relationships of size 1. However, the model completely fails for size 2 and 3. 

Perhaps even more surprisingly, the bottom half of the Table~\ref{reltype} shows that NTP does not scale with respect to the number of predicates \emph{even when} the relationship is at its simplest: as the total number of predicates increase the effective size of the state space and decrease the signal to noise ratio, the model performance decreases sharply. 

Our main conclusion is \textbf{NTP does not learn complicated relationships well}. Nonetheless, the model still
achieves better-than-random fact prediction accuracy. One possible reason is that the learning algorithm can place
predicates in a relationship close to each other in the embedding space and unifying the predicates directly - we
analyze this in the next section.

\subsection{Diagnosis}

Why does NTP perform so disappointingly in learning relations? We argue that the problem lies in the nature of its greedy optimization and corresponding lack of exploration.

The  model works in a winner-takes-all greedy fashion: (1) picks the highest scoring proof; (2) if a correct fact was
proved, increase the score	of the proof.  Such a process naturally leads to highly stochastic outcomes, where the
final winner depends on the structure that was initially chosen. See the example below:
\begin{example}{Example II: Failed rule learning}
\small 
	Consider again the relationship $\texttt{HasSibling}(x) \leftarrow \texttt{HasBrother}(x)$ and rule $B(x)
	\leftarrow A(x)$. We try to prove \texttt{HasSibling}(\textsc{Emily}), knowing
	\texttt{HasBrother}(\textsc{Emily}). There are two obvious candidate proofs. The desired proof applies the
	rule, unifying $A$ with \texttt{HasBrother} and $B$ with \texttt{HasSibling}. The alternative proof directly
	unifies \texttt{HasBrother} and \texttt{HasSibling}. If the direct unification proof starts out with a higher
	score (due to random initialization, as explained below), the rule embeddings will not be updated and the
	unification proof will have a higher score permanently.
\end{example}
In other words, the model sticks with the first somewhat reasonable proof, trapped in a stable local minimum,
rather than exploring for the best proof. If the winner-takes-all phenomenon is indeed causing problems for the
model, one would expect model performance to be highly dependent on initialization. Our experiments show that
is indeed the case. The standard deviation of recall for the model on relationships of size 1 and order 1,
purely from repeated initialization (calculated from 50 initializations each for 20 dataset draws) is equal to
0.39. The high variance reflects feast-or-famine results of the model: for advantageous initializations the
model finds the correct rule with close to perfect confidence, and for all other initializations the confidence of
the rule is close to zero.

\begin{table}
	\caption{Relationship learning performance with respect to initialization}
	\label{ratioinit}
	\centering
	{\small
	\begin{tabular}{ccccccc}
						 
		\multicolumn{2}{c}{Rule body} &   &\multicolumn{2}{c}{Rule performance}
		& \multicolumn{2}{c}{Fact performance} \\
		\cmidrule(lr){1-2}
		\cmidrule(lr){4-5}
		\cmidrule(lr){6-7}
		Size               & Order                  & r    & Recall & PR-AUC & MRR  & ROC-AUC \\
		\midrule
		\multirow{4}{*}{1} & \multirow{4}{*}{Unary} & 1.0  & 0.42   & 0.58   & 0.26 & 0.90    \\
		                   &                        & 0.9  & 0.76   & 0.86   & 0.28 & 0.94    \\
		                   &                        & 0.75 & 0.96   & 0.98   & 0.29 & 0.95    \\
		                   &                        & 0.5  & 1.0    & 1.0    & 0.29 & 0.95    \\
		\midrule
		\multirow{4}{*}{2} & \multirow{4}{*}{Unary} & 1.0  & 0.00   & 0.35   & 0.06 & 0.79    \\
		                   &                        & 0.9  & 0.06   & 0.37   & 0.07 & 0.80    \\
		                   &                        & 0.75 & 0.40   & 0.67   & 0.14 & 0.85    \\
		                   &                        & 0.5  & 0.90   & 0.98   & 0.26 & 0.95    \\
		\bottomrule
	\end{tabular}
	} \vskip -0.5em
\end{table}

Table~\ref{ratioinit} contains a more thorough investigation of the effect of initialization on performance. After initializing all embeddings, we determine the rule for which the rule predicate embeddings are closest to the \emph{ground-truth} relationship embeddings, and move those rule embeddings even closer, multiplying the distance by a ratio $r \leq 1$. Modest reductions in distance lead to sharply improved performance; a larger reduction of 0.5 increases recall from 0.0 to 0.9 for relationships of size 2. 

\begin{figure}[t]
	\centering
	\includegraphics[width=0.8\textwidth]{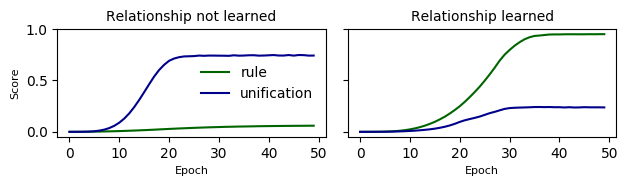}
	\caption{\small Development of rule and unification proof scores during training, conditional on whether or not relationship was successfully learned}
	\label{fig:ruleunification}
	\vskip -0.5em
\end{figure}

Initialization is clearly important, but is it a question of competing proofs? Define the \textit{rule score} as the highest relationship decoding score amongst all the rules, and the \textit{unification score} as the score of unifying the head and the closest body predicate directly. There exists a strongly negative (-0.85) correlation between the rule score and the unification score at the end of training. Figure \ref{fig:ruleunification} displays how the rule and unification scores develop during training, conditional on whether or not the correct relationship 
is eventually learned. The results show training runs go in one of two ways: either the rule catches on, and the rule score increases while the unification score increases only modestly, or it does not and the unification score increases while the rule score remains low. This pattern strongly suggests these two proof types are competing in the simple case of a relationship of size 1, order 1, and provide support for the winner-takes-all hypothesis.

% !TEX root = ms.tex
\section{Rule learning needs exploration}
\label{exploration}

This section proposes an adjusted version of the model in \cite{rocktaschel2017end}, propagating gradients to additional proofs beyond the best-scoring proof in order to encourage exploration and reduce the winner-takes-all property of the original model. Taking inspiration from beam search, we aim to keep around proofs with lower current scores that may prove successful later in training, when the influence of other data points has been
incorporated into the embeddings. 

Propagating gradients to more proofs is equivalent to altering the original loss function for a single training example (Eqn \ref{eqn:loss}) by summing losses over a set of proofs.
\begin{equation}
	\mathcal{L}_i(\bm{\theta}) = \sum_{k \in K}  -y_i \log(\rho_{ik}) - (1 - y_i) \log (1 - \rho_{ik} )
\end{equation}
Different choices of the set $K$ then lead to different heuristics. We employ two sets of heuristics for
exploration. Top-$k$ propagates gradients to the top-$k$ highest scoring proofs. Define a \emph{proof path} as a
sequence of categories (facts and different rule templates) of logical statements  applied in a proof. For example, with one rule template of size 2,
the categories are fact and rule, and the two proof paths of depth 1 are (fact) and (rule, fact, fact).

The \textsc{all-path} heuristic propagates gradients to the top-scoring proof from each high-level proof path to encourage varied exploration. 

Table~\ref{headline} demonstrates the benefits of incorporating exploration heuristics, especially in learning complicated relationships. Contrasting to what is in Table~\ref{reltype}, performances are consistently improved by both types of heuristics and their combination and translate into higher fact prediction accuracy.

\begin{table}[t]
	\caption{NTP performance by exploration heuristics}
	\label{headline}
	\centering
	{\small
	\begin{tabular}{ccccccc}
						 
		\multicolumn{2}{c}{Rule body} & &\multicolumn{2}{c}{Rule performance} & \multicolumn{2}{c}{Fact
		performance} \\
		\cmidrule{1-2}
		\cmidrule(lr){4-5}
		\cmidrule(lr){6-7}
		Size               & Order                   & Exploration heuristic & Recall        & PR-AUC        & MRR
		& ROC-AUC       \\
		\midrule
		\multirow{3}{*}{1} & \multirow{3}{*}{Unary}  & Top-$2$                 & 0.9           & 0.94          & 0.28
		                   & 0.93 \\
		                   &                         & \textsc{all-path}              & \textbf{1.0}  & \textbf{1.0}  & 0.29
		                   & 0.93 \\
		                   &                         & Top-$2$ \textsc{all-path}        & \textbf{1.0}  & \textbf{1.0}  &
		                   \textbf{0.29} & \textbf{0.95} \\
		\midrule
		\multirow{3}{*}{1} & \multirow{3}{*}{Binary} & Top-$2$                 & 0.88          & 0.93          & 0.52
		                   & 0.93 \\
		                   &                         & \textsc{all-path}              & \textbf{0.99} & 0.96          &
		                   \textbf{0.57} & 0.98 \\
		                   &                         & Top-$2$ \textsc{all-path}        & 0.97          & \textbf{0.99} & 0.54
		                   & 0.96 \\
		\midrule
		 \multirow{3}{*}{2} & \multirow{3}{*}{Unary}  & Top-$2$                 & 0.14          & 0.46          & 0.07
		                   & 0.84 \\
		                   &                         & \textsc{all-path}              & 0.14          & 0.46          & 0.07
		                   & 0.84 \\
		                   &                         & Top-$2$ \textsc{all-path}        & \textbf{0.92} & \textbf{0.95} &
		                   \textbf{0.27} & \textbf{0.96} \\ 
		\midrule
		\multirow{3}{*}{2} & \multirow{3}{*}{Binary} & Top-$2$                 & 0.16          & 0.49          & 0.21
		                   & 0.88 \\
		                   &                         & \textsc{all-path}              & 0.04          & 0.49          & 0.08
		                   & 0.89 \\
		                   &                         & Top-$2$ \textsc{all-path}        & \textbf{0.92} & \textbf{0.96} &
						\textbf{0.54} & \textbf{0.98} \\
		\midrule
		\multirow{4}{*}{3} & \multirow{4}{*}{Unary}    & Top-$2$                 & 0.08          & 0.52          &
							0.03 & 0.82 \\
							&                         & \textsc{all-path}              & 0.18          & 0.48          &
							0.07 & 0.85 \\
							&                         & Top-$2$ \textsc{all-path}        & 0.38 & 0.55 & 0.14 & 0.93 \\
							
							&                         & Top-$3$ \textsc{all-path}        & \textbf{0.64} & \textbf{0.73} &
							\textbf{0.20} & \textbf{0.96} \\						   							
		\bottomrule    
	\end{tabular}
	}
\end{table}

Table \ref{constants2} repeats the experiment from Section \ref{ablations} using the new exploration heuristic.
Here we find that the model can learn from extremely small amounts of data -- as few as 11 active facts.

\begin{table}
	\caption{NTP performance by number of constants using \textsc{all-path} top-$2$ heuristic}
	\label{constants2}
	\centering
	{\small
	\begin{tabular}{cccccccc}
						 
		\multicolumn{2}{c}{Rule body} & \multicolumn{3}{c}{Data size}
		&\multicolumn{2}{c}{Rule performance} & Fact Performance \\
		\cmidrule(lr){1-2}
		\cmidrule(lr){3-5}
		\cmidrule(lr){6-7}
		\cmidrule(lr){8-8}
		Size & Order & Constants & Total facts & Active facts & Recall & PR-AUC  & ROC-AUC \\
		\midrule
		\multirow{4}{*}{1}& \multirow{4}{*}{Unary} &  50  &  137  & 23 & 0.76 & 0.79 & 0.86  \\
		&& 100 & 274 & 47 & 0.96 & 0.98 & 0.92 \\
		&& 200 & 546 & 92 & 1.0 & 1.0 & 0.92  \\
		&& 800  & 2194 & 388 & 1.0 & 1.0 & 0.96 \\
		\midrule
		\multirow{4}{*}{2}& \multirow{4}{*}{Unary} &  50  & 133 & 11 & 0.16 & 0.35 & 0.80     \\
		&& 100 & 266 & 21 & 0.64 & 0.77 & 0.90 \\
		&& 200 & 532 & 43 & 0.90 & 0.94 & 0.93  \\
		&& 800 & 2139 & 185& 0.98 & 0.99 & 0.97 \\    
		\bottomrule
	\end{tabular}
	}
\end{table}

\section{Discussion}

Neural theorem proving is a promising combination of logical learning and neural network approaches. In this work,
we evaluate the performance of the NTP algorithm on synthetic logical datasets with injected relationships. We show
that NTP has difficulty recovering relationships in all but the simplest settings. Our experiments suggest the
problem lies in the presence of structural local minima, due to the winner-takes-all property of the model. We
alter the NTP algorithm to increase exploration, which sharply improves performance.

We believe there are several lesssons to be drawn from this work beyond the immediate application to NTP. First,
that it is helpful to look at synthetic data when evaluating prediction models that involve structure learning, as
final prediction accuracy can mask problems with the structure learning component. Second, that learning discrete
structures as an intermediate step can be accompanied by severe structural local minima, which can be avoided
through additional exploration.

\textbf{Acknowledgments}  {\small This work is partially supported by NSF Awards IIS-1513966/ 1632803/1833137, CCF-1139148, DARPA Award\#: FA8750-18-2-0117,  DARPA-D3M - Award UCB-00009528, Google Research Awards, gifts from Facebook and Netflix, and ARO\# W911NF-12-1-0241 and W911NF-15-1-0484. We thank Shariq Iqbal, Zhiyun Lu, and Bowen Zhang for helpful comments.}

\section*{Supplemental Parameter Experiments}
\begin{table}[!htbp]
	\caption{NTP base model performance by number of rules}
	\label{rules}
	\centering
	\begin{tabular}{ccccccc}
						 
		\multicolumn{2}{c}{Rule body} &   &\multicolumn{2}{c}{Rule performance}
		& \multicolumn{2}{c}{Fact performance} \\
		\cmidrule(lr){1-2}
		\cmidrule(lr){4-5}
		\cmidrule(lr){6-7}
		Size & Order & Rules & Recall & PR-AUC & MRR & ROC-AUC \\
		\midrule
		\multirow{5}{*}{1}&\multirow{5}{*}{Unary} & 3 & 0.62 &	0.76&	0.28 & 0.92  \\
		& &  5 &0.68 &	0.75&	0.28 &	0.93  \\
    & & 10 &0.76 &	0.78&	0.28 &	0.92   \\
    & & 20 & 0.86 &	0.83&	0.28 &	0.94  \\
    & & 50 & 0.9 &	0.84	&0.27 &	0.92  \\
		\midrule
		\multirow{5}{*}{1}&\multirow{5}{*}{Unary} & 3 &0.0 &	0.39	&0.04 &	0.81  \\
		& &  5 &0.04 &	0.3	& 0.04 &	0.82  \\
    & & 10 & 0.04 &	0.21	& 0.04 &	0.82   \\    
    & & 20 & 0.0 &	0.14	& 0.04 &	0.81   \\
    & & 50 &0.06 &	0.09	& 0.04 &	0.81   \\
		\bottomrule
	\end{tabular}
\end{table}

This section contains experiments verifying that the conclusions in the body of the paper hold broadly and are not sensitive to model parameters. 

\subsection*{Rules}

Table \ref{rules} shows how performance of the base model varies with the number of instantiated rules. Increasing the number of rules does improve relationship recall for relationships of size 1 and order 1, improving recall and precision. However, adding rules is not a panacea - for even slightly more complex relationships of size 2 and
order 1, increasing the number of rules improves recall only slightly at the cost of a large reduction in precision.

\subsection*{Rule Probability}

Table \ref{ruleprob} shows how performance of the base model and heuristic vary with the strength of the relationship, defined as the probability for the head predicate of a relationship to hold for a set of constants if 
the body predicates hold for that set of constants. The algorithm is still able to learn non-deterministic relationships, although rule learning and fact prediction performance do decrease as relationship strength decreases. The model with exploration heuristic still performs much better than the base algorithm with nondeterministic relationships. Note that the upper bound on fact prediction accuracy also decreases as the relationship strength decreases.

\begin{table}[t]
	\caption{NTP base model and exploration heuristic performance by relationship probability}
	\label{ruleprob}
	\centering
	\begin{tabular}{cccccccc}
						 
		\multicolumn{2}{c}{Rule body} &  & &\multicolumn{2}{c}{Rule performance}
		& \multicolumn{2}{c}{Fact performance} \\
		\cmidrule(lr){1-2}
		\cmidrule(lr){5-6}
		\cmidrule(lr){7-8}
		Size & Order &  Exploration & $p_r$  & Recall & PR-AUC & MRR & ROC-AUC \\
		\midrule
		\multirow{4}{*}{1}&\multirow{4}{*}{Unary} & \multirow{4}{*}{Vanilla} & 1.0 & 0.62 &0.76&	0.28 &	0.92  \\
		& &   & 0.9  & 0.36 &	0.51&	0.11 &	0.83\\
    & &  & 0.8 & 0.18 &	0.4&	0.07 &	0.80 \\
    & &  &  0.7& 0.16 &	0.37	&0.05 &	0.73 \\
    \midrule
    \multirow{4}{*}{1}&\multirow{4}{*}{Unary} & \multirow{4}{*}{Top-$2$ \textsc{all-path}} & 1.0 & 1.0 &	1 &	0.29 &	0.95   \\
		& &   & 0.9 & 0.98 &	0.99 &	0.12 &	0.87 \\
    & &  &  0.8& 0.9 & 	0.9 &	0.08 &	0.82  \\
    & &  &  0.7& 0.44 &	0.51	& 0.05 &	0.73  \\
    \midrule
		\multirow{4}{*}{2}&\multirow{4}{*}{Unary} & \multirow{4}{*}{Vanilla} & 1.0 & 0.02 &	0.39 &	0.04 &	0.81   \\
		& &   & 0.9 &  0.0 &	0.36 &	0.03 &	0.79 \\
    & &  & 0.8 &  0.04 &	0.36 &	0.03 &	0.77 \\
    & &  &  0.7& 0.04 &	0.39 &	0.02 &	0.74 \\
    \midrule
    \multirow{4}{*}{2}&\multirow{4}{*}{Unary} & \multirow{4}{*}{Top-$2$ \textsc{all-path}} & 1.0 &0.92 &	0.95 &	0.27 &	0.96
    \\
		& &   &  0.9& 0.96 &	0.98 &	0.12 &	0.95 \\
    & &  &  0.8& 0.82 &	0.88 &	0.07 &	0.91  \\
    & &  &  0.7& 0.34 &	0.44 &	0.04 &	0.82  \\
		\bottomrule
	\end{tabular}
\end{table}

\subsection*{Relationships}

Table \ref{relationships} shows the effect of injecting a second relationship of the same type on performance of
the base model as well as the model with the Top-2-all-type exploration heuristic. The additional relationship
leads to a sharp reduction in relationship learning performance, though the model with exploration heuristic still
performs much better than the base algorithm at learning multiple relationships. 

\begin{table}[t]
	\caption{NTP base model and exploration heuristic performance by number of relationships}
	\label{relationships}
	\centering
	\begin{tabular}{cccccccc}
						 
		\multicolumn{2}{c}{Rule body} &  & &\multicolumn{2}{c}{Rule performance}
		& \multicolumn{2}{c}{Fact performance} \\
		\cmidrule(lr){1-2}
		\cmidrule(lr){5-6}
		\cmidrule(lr){7-8}
		Size & Order &  Exploration & Relationships  & Recall & PR-AUC & MRR & ROC-AUC \\
		\midrule
		\multirow{2}{*}{1}&\multirow{2}{*}{Unary} & \multirow{2}{*}{Vanilla} & 1 & 0.62 &	0.76 &	0.28 &	0.92   \\
		& &   & 2  &0.38 &	0.61 &	0.31 &	0.87 \\
    \midrule
    \multirow{2}{*}{1}&\multirow{2}{*}{Unary} & \multirow{2}{*}{Top-$2$ \textsc{all-path}} & 1& 1.0 &	1 &	0.29 & 	0.95 \\
		& &   & 2 & 0.63 &	0.71	& 0.41 &	0.96 \\
    \midrule
		\multirow{2}{*}{2}&\multirow{2}{*}{Unary} & \multirow{2}{*}{Vanilla} & 1 & 0.02 &	0.39 &	0.04 &	0.81  \\
		& &   & 2 & 0.01 &	0.48 &	0.04 &	0.80 \\

    \midrule
    \multirow{2}{*}{2}&\multirow{2}{*}{Unary} & \multirow{2}{*}{Top-$2$ \textsc{all-path}} & 1 & 0.92 &	0.95 &	0.27 &	0.96  \\
		& &   &  2& 0.47 &	0.7	 & 0.26 &	0.93 \\
		\bottomrule
	\end{tabular}
\end{table}

\subsection*{$K_{max}$ heuristic}

Following \citet{rocktaschel2017end}, for proofs that unify with several different facts, we only retain the top-k
highest scoring fact unifications per fact in each branch of the proof tree to reduce computational demands. For
example, for a proof path of type (rule, fact, fact), we do not take the maximum over all $n_r \times n_f \times
n_f$ proofs, but only $n_r \times top-k \times n_f$ proofs, where we retain the $n_r \times top-k$ highest fact
unification scores in the second step. 

Table \ref{kmax} shows that varying this $K_{max}$ parameter has a minimal effect on the outcome of the algorithm. 

\begin{table}[t]
	\caption{NTP base model and exploration heuristic performance by $K_{max}$ value}
	\label{kmax}
	\centering
	\begin{tabular}{cccccccc}
						 
		\multicolumn{2}{c}{Rule body} &  & &\multicolumn{2}{c}{Rule performance}
		& \multicolumn{2}{c}{Fact performance} \\
		\cmidrule(lr){1-2}
		\cmidrule(lr){5-6}
		\cmidrule(lr){7-8}
    Size & Order &  Exploration & $K_{max}$  & Recall & PR-AUC & MRR & ROC-AUC \\
    \midrule
		\multirow{3}{*}{2}&\multirow{3}{*}{Unary} & \multirow{3}{*}{Vanilla} & 10 & 0.02 &	0.39 &	0.04 &	0.81  \\
    & &   & 20 & 0.02 &	0.39 &	0.04 &	0.81 \\
    & &   & $\infty$ &  0.02 &	0.39 &	0.04 &	0.81 \\

    \midrule
    \multirow{3}{*}{2}&\multirow{3}{*}{Unary} & \multirow{3}{*}{Top-$2$ \textsc{all-path} } & 10 & 0.92 &	0.95 &	0.27 &	0.96  \\
    & &   &  20&  0.88 &	0.93 &	0.26 &	0.96 \\
    & &   & $\infty$ & 0.88 &	0.93 &	0.26 &	0.96 \\
		\bottomrule
	\end{tabular}
\end{table}

\bibliography{ntp}

\bibliography{ntp}

\end{document}